\documentclass{article}
\usepackage{arxiv}
\usepackage[utf8]{inputenc}
\usepackage[T1]{fontenc}
\usepackage{microtype}
\usepackage{amsmath,amssymb,booktabs,tabularx,graphicx,multirow}
\usepackage[numbers,sort&compress]{natbib}
\usepackage{needspace,float,placeins}
\usepackage{pdflscape}
\usepackage[unicode,hidelinks]{hyperref}
\hypersetup{
pdftitle={Graph learning-assisted logic-based Benders decomposition for distributionally robust in-station train rescheduling under disruptions},
pdflang={en-US},
pdfauthor={},
pdfsubject={},
pdfkeywords={in-station train rescheduling, disruptions, uncertain closure duration, distributionally robust optimization, logic-based Benders decomposition, graph neural networks}
}

\newcolumntype{C}[1]{>{\hsize=#1\hsize\linewidth=\hsize\centering\arraybackslash}X}

\newcommand{\tableformat}{\centering\small\setlength{\tabcolsep}{5pt}\renewcommand{\arraystretch}{1.0}}
\graphicspath{{figures/}}
\title{Graph learning-assisted logic-based Benders decomposition for distributionally robust in-station train rescheduling under disruptions}
\author{
Xiaobin Li$^*$ \\
  School of Transportation Engineering \\
  East China Jiaotong University\\
  Nanchang, Jiangxi 330038, China   \\
  \texttt{Yuncifor@outlook.com} \\
\And
 Yanbin Gao\\
  School of Transportation Engineering \\
East China Jiaotong University\\
  Nanchang, Jiangxi 330038, China   \\
  \And
   Jianing Long \\
  School of Transportation Engineering \\
East China Jiaotong University\\
  Nanchang, Jiangxi 330038, China   \\
\And
 Wuming Lei \\
  School of Transportation Engineering \\
East China Jiaotong University\\
  Nanchang, Jiangxi 330038, China   \\
}
\date{}
\renewcommand{\undertitle}{\mbox{}}
\begin{document}
\maketitle
\begin{abstract}
Resource closures, arrival delays, and processing-time extensions jointly alter station operations during disruptions. Uncertain reopening times further complicate decisions about holding trains and selecting alternative routes. We use observations of disruption-resolution stages to update conditional predictions of the remaining closure duration. A distributionally robust model accounts for errors in scenario probability estimates and jointly optimizes train routes, sequences, and timings. Logic-based Benders decomposition coordinates routing and resource scheduling. Conditional cuts exploit resource-use patterns, and a graph neural network guides the solution order of resource relaxations. Together, these components form the GNN-LBBD algorithm. Across 28 test instances for a five-track station, GNN-LBBD achieved a mean total completion delay of 61.13 train-min. This was 16.63\% lower than tabu search and 20.06\% lower than variable neighborhood search. Its mean total computation time was close to that of tabu search. On six representative instances, using stage observations reduced mean total completion delay by approximately 21.7\%. Continued updating of reopening-time predictions allows route and entry-time adjustments during closures, reducing delays caused by waiting for reopening.The implementation is available at \url{https://github.com/yuncifor/RESCHEDULING-UNDER-DISRUPTIONS}.
\end{abstract}

\section{Introduction}
Train operations within a station require coordinated use of platform tracks and station throats \citep{lusby2011}. Resource closures restrict capacity, arrival delays alter operation timing, and processing-time extensions increase resource occupancy. Together, these disturbances may intensify resource conflicts and cause subsequent trains to wait and incur delays \citep{cacchiani2014}. Field observations obtained during a closure also change dispatchers' estimates of the reopening time. This study examines how these observations can inform train holding, route choice, and sequencing to reduce total completion delay.

Joint routing and timing optimization integrates resource selection and occupancy timing into a unified schedule. Carey and Carville \citep{carey2003} studied train scheduling and platform assignment at busy stations. Pellegrini et al.\ \citep{pellegrini2014} jointly optimized disrupted train routes and schedules using a detailed infrastructure representation. However, when the reopening time is specified as an input, schedule evaluation depends on the assumed resource-availability intervals. New observations may revise reopening-time estimates and change the expected losses and ranking of candidate schedules. This connects prediction updates with subsequent scheduling adjustments.

Uncertainty in reopening times makes schedule comparisons depend on both possible resource-availability times and their probabilities. Corman and Kecman \citep{corman2018} used Bayesian networks to update train delay distributions as new information became available. Ben-Tal et al.\ \citep{bental2013} used probability uncertainty sets to account for errors in estimated probabilities. Yang et al.\ \citep{yang2026} developed a distributionally robust model for joint timetable and maintenance-window rescheduling under uncertain disruption duration. Within a station, different routes and operation sequences yield different completion delays across scenarios. Changes in probability estimates may therefore alter the loss ranking of candidate schedules. Routing, sequencing, and timing optimization should account for both conditional predictions from field observations and errors in probability estimates.

Discrete route choices and multiresource occupancy constraints couple scheduling decisions across reopening scenarios. El\c{c}i and Hooker \citep{elci2022} used logic-based Benders decomposition to coordinate discrete decisions and scenario losses through master and scheduling subproblems. Varga et al.\ \citep{varga2024} used graph neural networks to assist cut strengthening and reduce the required number of subproblem solves. Cuts conditioned on complete route combinations restrict scheduling lower bounds to the corresponding combinations. Different combinations may nevertheless share train--resource use patterns. These shared patterns provide a basis for extending the applicability of scheduling lower bounds. Resource relaxations also differ in their lower-bound gains and solution times. Their solution order therefore affects lower-bound strengthening within a limited computation budget.

We develop a distributionally robust in-station train rescheduling model that updates reopening-time predictions with field observations. GNN-LBBD solves the model using graph learning-assisted logic-based Benders decomposition. The main contributions are as follows:
\begin{enumerate}
\renewcommand{\labelenumi}{\arabic{enumi})}
\item We integrate conditional predictions of the remaining closure duration with train routing, sequencing, and timing. A scenario probability ambiguity set accounts for probability estimation errors. Decisions remain consistent across scenarios with identical observation histories.
\item We construct conditional cuts from train--resource use patterns. Single-resource scheduling lower bounds remain applicable to different route combinations satisfying these conditions. A graph neural network learns lower-bound gains per unit of solution time to guide the solution order of resource relaxations.
\item Prediction comparisons distinguish the effects of using observations from those of the prediction representation. Execution records link revised reopening-time estimates to route choices, entry times, and subsequent delays. Further experiments examine scheduling performance under closure-duration prediction errors.
\end{enumerate}

\section{Related work}
\subsection{In-station train routing and timing optimization}
\label{sec:station-literature}
In-station train scheduling models differ in how closely they integrate track assignment, route choice, and timing. Zwaneveld et al.\ \citep{zwaneveld1996} formulated in-station routing as a node-packing problem. Billionnet \citep{billionnet2003} used integer programming to assign trains to available tracks. Caprara et al.\ \citep{caprara2011} incorporated route and timing adjustments into track assignment. Caimi et al.\ \citep{caimi2011} integrated conflict-free routing and timing in a resource-constrained multicommodity flow model. Scheduling adjustments therefore concern both resource selection and the timing of resource occupancy.

The granularity of the resource representation further determines which conflicts a model can identify. Rodriguez \citep{rodriguez2007} used constraint programming for real-time train scheduling at a railway junction. Kumar et al.\ \citep{kumar2018} applied constraint programming to scheduling at large terminal stations. Cardellini et al.\ \citep{cardellini2021} described in-station train operations using automated planning. Schutt et al.\ \citep{schutt2025} represented train movements and stops through section-occupancy constraints and compared constraint models and search strategies. Section-level representations capture shared resources and their release order along arrival and departure routes. Train entry times are consequently constrained by both platform-track and throat resources.

\subsection{Train rescheduling under uncertainty}
Uncertainty in disruption duration, running times, and dwell times affects both feasible train schedules and their associated losses. Zhan et al.\ \citep{zhan2024} systematically reviewed uncertainty in train timetable rescheduling and the methods used to address it. Zhu and Goverde \citep{zhu2020} considered uncertain disruption duration in a rolling horizon. They updated scenarios using predictions of the disruption end time and adjusted timetables through two-stage stochastic programming. Changes in future resource availability can thus inform subsequent schedules, rather than leaving the initial duration prediction unchanged.

Scenario probabilities, as well as scenario values, affect schedule evaluation. Yang et al.\ studied the joint rescheduling of train timetables and maintenance windows under uncertain disruption duration. They developed a rolling-horizon distributionally robust model and solved it with a generalized branch-and-Benders-cut algorithm. Their decisions cover train operations and maintenance arrangements. The uncertainty treatment concerns both disruption duration and scenario distributions. Conditional prediction updates revise future resource availability, whereas a probability ambiguity set accounts for distribution estimation errors. The two components address prediction information and scheduling loss evaluation, respectively.

\subsection{Logic-based Benders decomposition and learning-assisted optimization}
The coupling of routing and resource scheduling has motivated several solution approaches. Corman et al.\ \citep{corman2010} combined rerouting neighborhoods and a scheduling algorithm within a tabu search framework. Sam\`a et al.\ \citep{sama2017} used variable neighborhood search to coordinate routes and timings for disrupted trains. Pellegrini et al.\ \citep{pellegrini2015} developed RECIFE-MILP, a mixed-integer programming heuristic for real-time railway traffic management under detailed infrastructure constraints. These methods organize joint routing and scheduling adjustments through neighborhood search or mathematical programming.

Decomposition methods coordinate discrete choices and resource scheduling through feedback between a master problem and subproblems. Hooker \citep{hooker2007} established a logic-based Benders framework that combines mixed-integer programming and constraint programming. Lamorgese and Mannino \citep{lamorgese2015} exploited railway problem structure to develop an exact decomposition method for real-time dispatching. Han et al.\ \citep{han2025} combined mixed-integer programming and constraint programming within logic-based Benders decomposition for integrated train formation and resource scheduling in marshalling yards. To reduce repeated solution effort, Leutwiler et al.\ \citep{leutwiler2023} selected precomputed cut pools using delay similarity. Their method reuses solution information from similar instances while balancing the number of cuts against model size.

Graph learning can also directly generate scheduling decisions. Zhang et al.\ \citep{zhang2026} modeled train platforming and rescheduling as a Markov decision process and used a graph neural network policy to select scheduling actions. Their learning target is a scheduling policy. In contrast, learning-assisted cut strengthening predicts task subsets for constructing decomposition cuts. The two approaches act on scheduling decisions and the optimization process, respectively.

\section{Rescheduling model and solution method}
\subsection{Problem description and notation}
\begin{figure}[H]\centering
\includegraphics[width=0.86\linewidth]{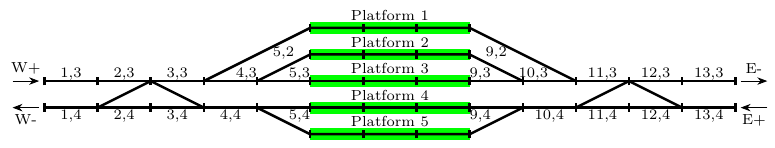}
\caption{Station layout}
\label{fig:topology}
\end{figure}

The station layout is taken from the public benchmark of Schutt et al., as shown in Fig.~\ref{fig:topology}. It comprises two station throats and five platform tracks. Trains enter and leave the platform tracks through the throats. The station has 45 controlled resources: four boundary sections, 26 internal sections, and 15 platform sections. Each platform track contains three platform sections.

\begin{figure}[!htbp]\centering
\includegraphics[width=\linewidth]{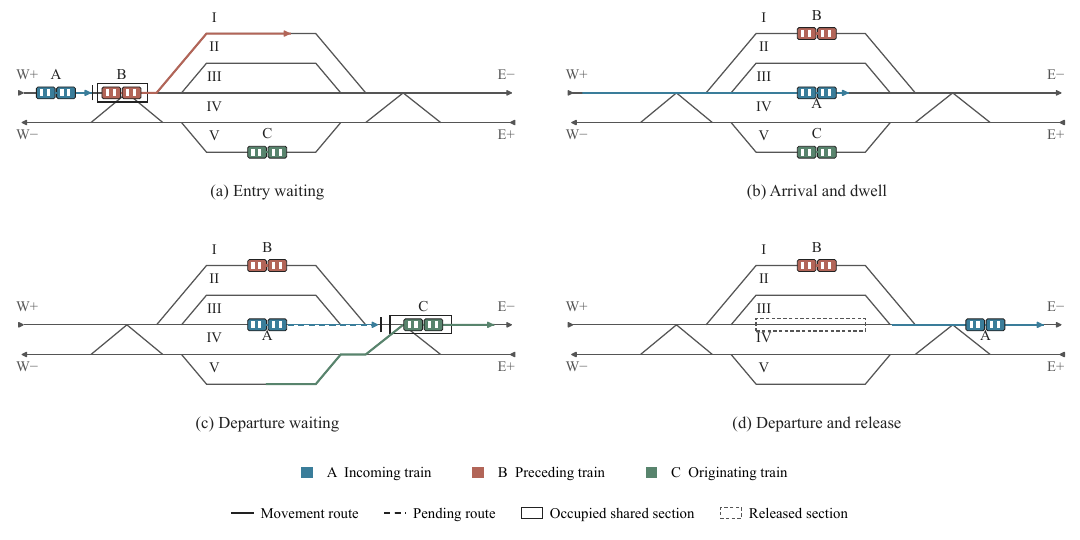}
\caption{In-station train operations}
\label{fig:dispatchprocess}
\end{figure}

Figure~\ref{fig:dispatchprocess} illustrates four operational stages: entry waiting, arrival and dwell, departure waiting, and departure with resource release. Trains occupy and release sections sequentially along their selected routes. Waiting occurs at the entry or on a platform track. Routes and movement sequences must be coordinated when trains share throat or platform-track resources. Following a disruption, rescheduling adjusts routes, sequences, and timings.

\Needspace{5\baselineskip}
Let $I$ denote the set of trains, $P_i$ the set of feasible routes for train $i$, and $\mathcal R$ the set of controlled resources. The planned completion time of train $i$ in the reference schedule $P_0$ is $C_i^0$. Table~\ref{tab:notation} lists the main notation. All time quantities in the model are expressed in seconds.

\begin{table}[!htb]\tableformat
\caption{Main notation}
\label{tab:notation}
\begin{tabularx}{\linewidth}{@{}C{0.625}C{1.375}@{}}\toprule
Symbol & Definition\\\midrule
$I,\ I_k^{\mathrm u}$ & Set of all trains; set of trains not yet started at update $k$\\
$P_i,\ B_{ip}$ & Feasible routes for train $i$; ordered sections of route $p$\\
$\mathcal R,\ r$ & Set of controlled resources and resource index\\
$t_k,\ \mathcal H_k$ & Time of update $k$ and information history available at that time\\
$R_k$ & Remaining closure duration at update $k$\\
$\Omega_k,\widehat p_{k\omega}$ & Conditional scenario set and nominal probability of scenario $\omega$\\
$C_i^0$ & Completion time of train $i$ in the reference schedule $P_0$\\
$\Delta_i^{A},\Delta_i^{D}$ & Exogenous arrival delay and additional required processing time\\
$x_{ipn},\ h_{in}$ & Binary variables for starting at node $n$ on route $p$, and for starting at node $n$\\
$s_{ib\omega},\ f_{ib\omega}$ & Start and release times of train $i$ on section $b$ in scenario $\omega$\\
$d_{i\omega}$ & Dwell and waiting duration of train $i$ in scenario $\omega$\\
$C_{i\omega},\ D_{i\omega}$ & Completion time and completion delay of train $i$ in scenario $\omega$\\
$\pi,\ L_\omega(\pi)$ & Scheduling policy and its total completion delay in scenario $\omega$\\
$\rho,\mathcal P_k$ & Scenario probability deviation radius and probability ambiguity set\\
\bottomrule\end{tabularx}
\end{table}

\subsection{Distributionally robust train rescheduling model}
\label{sec:recovery-model}
\subsubsection{Uncertainty and information structure}
The waiting time for the original route remains uncertain during a closure. Rerouting may increase competition for other resources. Dispatchers update remaining closure-duration estimates using actual train states, observed disturbances, and the disruption-resolution stage. Observations of this stage are subject to delivery delays and misclassification. Bayesian updating combines stage-transition and observation models.

Let $\mathcal H_k$ denote the information available at time $t_k$ and $R_k$ the remaining closure duration. Its conditional distribution is:
\begin{equation}
F_k(u)=\Pr(R_k\le u\mid\mathcal H_k).
\label{eq:conditional-recovery}
\end{equation}
The conditional scenario set $\Omega_k$ and nominal probabilities $\widehat p_{k\omega}$ are constructed from $F_k$. Let $R_{k\omega}$ be the remaining closure duration in scenario $\omega$. The reopening time of a currently closed resource is $b_{r\omega}=t_k+R_{k\omega}$. Train routes, sequences, and timings are jointly optimized under these resource-availability conditions.

Each scenario also contains subsequent stage observations. Scenarios with identical observation histories share an information node $n\in\mathcal N_k$. Node $n$ corresponds to the interval $[\ell_n,u_n)$, and $\operatorname{Path}(\omega)$ denotes the nodes visited by scenario $\omega$.

Decisions can depend only on the information available at the time. Let $\sigma_{\omega\omega'}$ be the first time at which observations distinguish scenarios $\omega$ and $\omega'$. Set $\sigma_{\omega\omega'}=\infty$ if they remain indistinguishable. Let $s_{i\omega}^{\mathrm{start}}$ be the train start time, $a_{ip\omega}$ indicate the use of route $p$, and $\zeta_{ip\omega}$ be the departure start time on that route. Nonanticipativity requires:
\begin{equation}
\begin{aligned}
\min\{s_{i\omega}^{\mathrm{start}},s_{i\omega'}^{\mathrm{start}}\}
<\sigma_{\omega\omega'}
&\ \Longrightarrow\
\begin{cases}
s_{i\omega}^{\mathrm{start}}=s_{i\omega'}^{\mathrm{start}},\\
a_{ip\omega}=a_{ip\omega'},\quad p\in P_i;
\end{cases}\\
a_{ip\omega}=a_{ip\omega'}=1,\quad
\min\{\zeta_{ip\omega},\zeta_{ip\omega'}\}
<\sigma_{\omega\omega'}
&\ \Longrightarrow\
\zeta_{ip\omega}=\zeta_{ip\omega'}.
\end{aligned}
\label{eq:nonanticipativity}
\end{equation}
A scheduling policy satisfying these nonanticipativity constraints is denoted by $\pi$.

\subsubsection{Objective and constraints}
Reopening-time predictions specify possible resource-availability times and their nominal probabilities. Errors in probability estimates alter the evaluation of candidate schedules. To account for probability estimation errors, we construct an $L_1$ ambiguity set around the nominal probabilities $\widehat p_k$:
\begin{equation}
\mathcal P_k=\left\{p\ge0:\boldsymbol1^\top p=1,\quad
\|p-\widehat p_k\|_1\le\rho\right\},\qquad 0\le\rho\le2.
\label{eq:ambiguity}
\end{equation}
Train completion delay and scenario loss are defined as:
\begin{equation}
D_{i\omega}=(C_{i\omega}-C_i^0)^+,\qquad
L_\omega(\pi)=\sum_{i\in I}D_{i\omega},\qquad (x)^+=\max\{0,x\}.
\label{eq:loss}
\end{equation}
Routes and operation sequences yield different completion delays across reopening scenarios. Each candidate policy is evaluated by its worst-case expected total completion delay over the probability ambiguity set. We select the policy that minimizes this loss:
\begin{equation}
J_k^*=\min_{\pi\in\Pi(\mathcal H_k)}
\ \max_{p\in\mathcal P_k}\sum_{\omega\in\Omega_k}p_\omega L_\omega(\pi).
\label{eq:objective}
\end{equation}
The feasible policy set $\Pi(\mathcal H_k)$ is defined by train-operation constraints, resource-use constraints, and the nonanticipativity constraints in Eq.~\eqref{eq:nonanticipativity}.

\textbf{Route and start-node selection.}
For each train $i\in I_k^{\mathrm u}$ that has not yet started, the route and start-node variables $x_{ipn},h_{in}\in\{0,1\}$ satisfy:
\begin{equation}
\begin{aligned}
\sum_{n\in\operatorname{Path}(\omega)}h_{in}&=1,\qquad
\sum_{p\in P_i}x_{ipn}=h_{in},\\
a_{ip\omega}&=\sum_{n\in\operatorname{Path}(\omega)}x_{ipn},\\
\sum_{n\in\operatorname{Path}(\omega)}\ell_n h_{in}
&\le s_{i\omega}^{\mathrm{start}}\le
\sum_{n\in\operatorname{Path}(\omega)}(u_n-1)h_{in},
\quad\omega\in\Omega_k.
\end{aligned}
\label{eq:choice}
\end{equation}

\textbf{Operation timing and section continuity.}
The original earliest ready time of train $i$ is $e_i^0$. Its disrupted ready time is $e_i=e_i^0+\Delta_i^A$. Let $\underline d_{ip}$ denote the minimum dwell time on route $p$ under normal conditions. The disrupted minimum processing time $w_{ip}$ satisfies:
\begin{equation}
\begin{aligned}
s_{i\omega}^{\mathrm{start}}&\ge\max\{t_k,e_i\},
&&i\in I_k^{\mathrm u},\\
w_{ip}&=
\begin{cases}
0,&i\text{ is an originating train},\\
\underline d_{ip}+\Delta_i^D,&i\text{ is a non-originating train},
\end{cases}\\
d_{i\omega}&\ge\sum_{p\in P_i}w_{ip}a_{ip\omega},
&&d_{i\omega}=0\quad\text{if }i\text{ is an originating train}.
\end{aligned}
\label{eq:ready-dwell}
\end{equation}
For the selected route $p$, let $b_1$ be its first section and $b^+$ the successor of section $b$. Parameters $\tau_b$ and $o_b$ denote the required section time and transition offset, respectively. The indicator $u_b$ identifies the dwell section, and $v_{ip}$ is the fixed running time of the route. Section occupancy and train completion times satisfy:
\begin{equation}
\begin{aligned}
s_{i b_1\omega}&=s_{i\omega}^{\mathrm{start}},\\
f_{ib\omega}&=s_{ib\omega}+\tau_b+u_bd_{i\omega},
&&b\in B_{ip},\\
s_{i,b^+,\omega}&=f_{ib\omega}+o_{b^+},\\
C_{i\omega}&=s_{i\omega}^{\mathrm{start}}+v_{ip}+d_{i\omega}.
\end{aligned}
\label{eq:block-link}
\end{equation}
The departure start time $\zeta_{ip\omega}$ is the start time of the first departure section after the dwell section. Time variables take integer-second values. Section times lie within the planning interval $[t^-,H]$, where $t^-$ is the batch start time and $H$ is the planning horizon.

\textbf{Resource occupancy and closures.}
Let $r(b)$ denote the resource associated with section $b$. The reservation endpoints for a section on the selected route are $\widetilde s_{ib\omega}$ and $\widetilde f_{ib\omega}$. They normally equal $s_{ib\omega}$ and $f_{ib\omega}$. For platform occupancy, the left endpoint is $t^-$ for originating trains, and the right endpoint is $H$ for terminating trains. Resource reservations and mutual exclusion satisfy:
\begin{equation}
\begin{aligned}
\mathcal I_{ir\omega}
&=\bigcup_{\substack{b\in B_{ip}\\r(b)=r}}
[\widetilde s_{ib\omega},\widetilde f_{ib\omega}),
\qquad a_{ip\omega}=1,\\
\mathcal I_{ir\omega}\cap\mathcal I_{jr\omega}
&=\varnothing,\qquad i\ne j,\quad r\in\mathcal R.
\end{aligned}
\label{eq:resource-intervals}
\end{equation}
If resource $r$ is closed over $[a_r,b_{r\omega})$, entry to the resource is restricted by:
\begin{equation}
s_{ib\omega}\le a_r-1\quad\text{or}\quad
s_{ib\omega}\ge b_{r\omega},\qquad r(b)=r.
\label{eq:resource-closure}
\end{equation}
Closure constraints apply to sections entering closed resources along the selected route. Platform occupancy that has already begun for originating trains is retained. Let $\mathcal R_{ip}^{\mathrm{in}}$ and $\mathcal R_{ip}^{\mathrm{out}}$ denote the resources entered during the continuous arrival and departure movements, respectively. For a closure with $t_k\in[a_r,b_{r\omega})$, the corresponding movements that have not yet started satisfy:
\begin{equation}
\begin{aligned}
a_{ip\omega}=1&\ \Longrightarrow\
s_{i\omega}^{\mathrm{start}}\ge b_{r\omega},
&&r\in\mathcal R_{ip}^{\mathrm{in}},\quad i\in I_k^{\mathrm u},\\
a_{ip\omega}=1&\ \Longrightarrow\
\zeta_{ip\omega}\ge b_{r\omega},
&&r\in\mathcal R_{ip}^{\mathrm{out}}.
\end{aligned}
\label{eq:journey-closure}
\end{equation}
Consecutive trains $(i,j)\in\mathcal E$ at the same entry retain their prescribed order:
\begin{equation}
s_{i\omega}^{\mathrm{start}}\le s_{j\omega}^{\mathrm{start}},
\qquad (i,j)\in\mathcal E,\quad\omega\in\Omega_k.
\label{eq:entry-order}
\end{equation}

\textbf{Executed operations.}
Let $I_k^{\mathrm s}$ be the set of trains that have started. Sets $\mathcal J_k^{\mathrm{end}}$ and $\mathcal J_k^{\mathrm{occ}}$ contain completed and ongoing section operations, respectively. The superscript $\mathrm{act}$ denotes actual routes, start times, and section times. Confirmed routes remain unchanged. All scenarios retain:
\begin{equation}
\begin{aligned}
a_{i,p_i^{\mathrm{act}},\omega}&=1,\qquad
s_{i\omega}^{\mathrm{start}}=s_i^{\mathrm{act}},
&&i\in I_k^{\mathrm s},\\
s_{ib\omega}&=s_{ib}^{\mathrm{act}},\qquad
f_{ib\omega}=f_{ib}^{\mathrm{act}},
&&(i,b)\in\mathcal J_k^{\mathrm{end}},\\
s_{ib\omega}&=s_{ib}^{\mathrm{act}},\qquad
f_{ib\omega}\ge t_k+1,
&&(i,b)\in\mathcal J_k^{\mathrm{occ}}.
\end{aligned}
\label{eq:executed}
\end{equation}

\subsection{Graph learning-assisted logic-based Benders decomposition}
\subsubsection{Master problem and joint scheduling subproblem}
\begin{figure}[!htbp]\centering
\includegraphics[width=\linewidth,height=0.38\textheight,keepaspectratio]{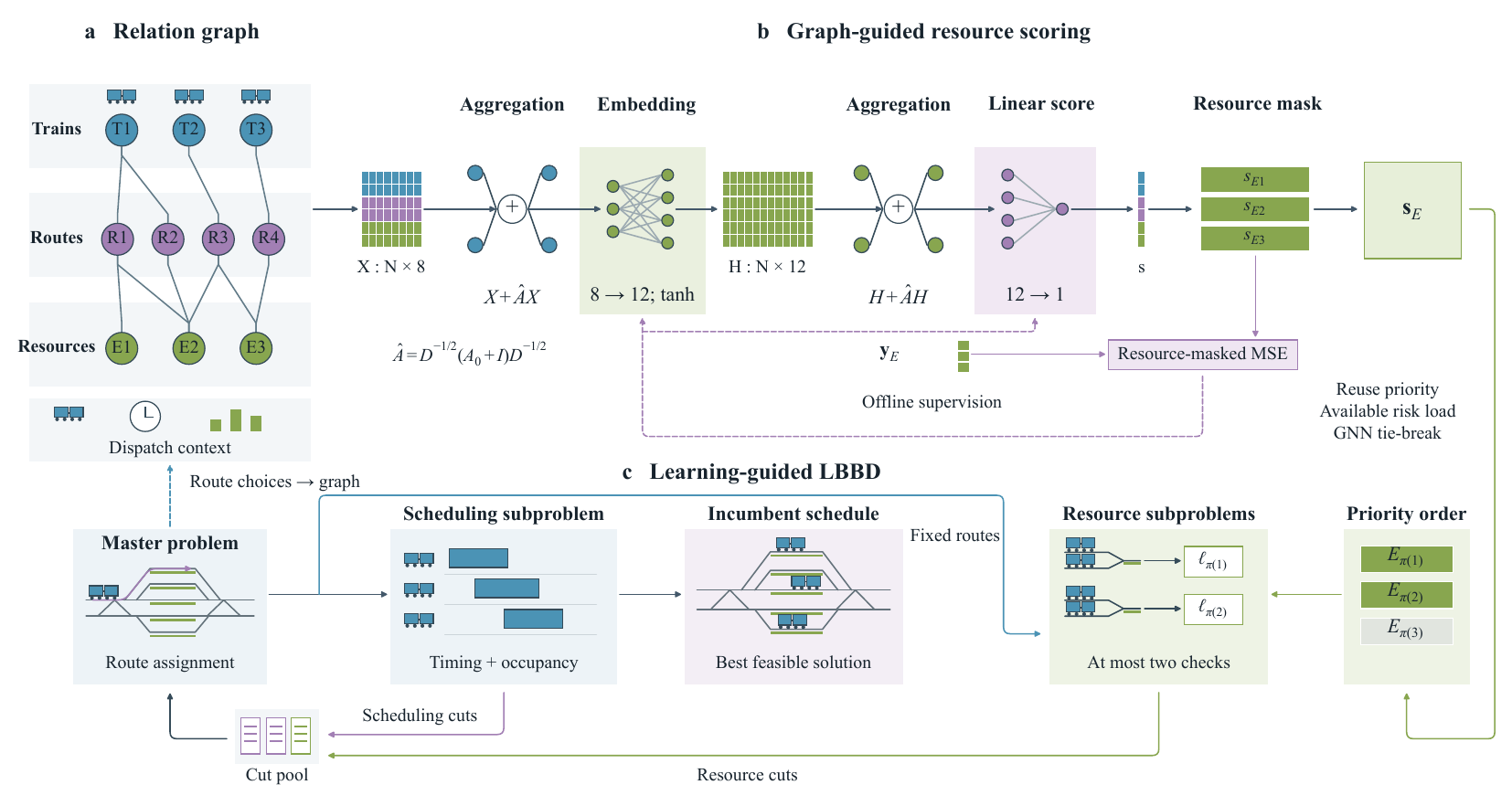}
\caption{GNN-LBBD framework}
\label{fig:architecture}
\end{figure}

Logic-based Benders decomposition coordinates routing and resource scheduling. Resource lower bounds and conditional cuts strengthen the master problem, while graph learning guides resource ordering. The algorithm is denoted by GNN-LBBD, and its framework is shown in Fig.~\ref{fig:architecture}.

Let $b=(x_{ipn})\in\mathcal B_k$. The earliest completion times $\underline C_{ipn\omega}$, obtained without resource competition, give the initial scenario lower bound:
\begin{equation}
\underline L_\omega^0(b)=
\sum_{i\in I_k^{\mathrm u}}
\sum_{n\in\operatorname{Path}(\omega)}
\sum_{p\in P_i}
(\underline C_{ipn\omega}-C_i^0)^+x_{ipn}.
\label{eq:initial-bound}
\end{equation}
The set $\mathcal B_k$ retains necessary conditions on route choice, start-time windows, and entry order. The calculation of $\underline C_{ipn\omega}$ retains ready times, minimum processing times, and route closure conditions. Choices with no feasible start-time window are removed from $\mathcal B_k$. Let $\underline s_{ipn\omega}$ be the corresponding earliest start time. A necessary condition for the prescribed order at the same entry is:
\begin{equation}
\sum_{\substack{n\in\operatorname{Path}(\omega)\\p\in P_i}}
\underline s_{ipn\omega}x_{ipn}
\le
\sum_{\substack{n\in\operatorname{Path}(\omega)\\p\in P_j}}
(u_n-1)x_{jpn},
\qquad (i,j)\in\mathcal E.
\label{eq:master-entry}
\end{equation}
For trains that have started, the corresponding start-time endpoints take their actual values.

Let $\eta_\omega$ and $\theta$ be lower-bound variables for scenario loss and the robust objective, respectively. Given a set of probability vectors $\mathcal V_k\subseteq\mathcal P_k$ and a set of Benders cuts $\mathcal C_k$, the master problem is:
\begin{equation}
\begin{aligned}
\min_{b,\eta,\theta}\quad&\theta\\
\text{s.t.}\quad&
b\in\mathcal B_k,\qquad
\eta_\omega\ge\underline L_\omega^0(b),\\
&\theta\ge\sum_{\omega\in\Omega_k}p_\omega\eta_\omega,
\qquad p\in\mathcal V_k,\\
&(b,\eta,\theta)\text{ satisfies }\mathcal C_k,\qquad
\eta\ge0,\quad\theta\ge0.
\end{aligned}
\label{eq:master}
\end{equation}
Initially, $\mathcal V_k=\{\widehat p_k\}$. After obtaining a candidate solution, we solve:
\begin{equation}
p^*\in\arg\max_{p\in\mathcal P_k}p^\top\eta.
\label{eq:probability-separation}
\end{equation}
If $(p^*)^\top\eta>\theta$, we add $p^*$ to $\mathcal V_k$ and resolve the master problem.

Let $\bar a_{ip\omega}=\sum_{n\in\operatorname{Path}(\omega)}\bar x_{ipn}$ denote the candidate routes. The joint subproblem fixes the route in each scenario and optimizes start intervals, resource-use sequences, and operation times:
\begin{equation}
Q_k(\bar a)=
\min_{\substack{\pi\in\Pi(\mathcal H_k)\\a(\pi)=\bar a}}
\ \max_{p\in\mathcal P_k}
\sum_{\omega\in\Omega_k}p_\omega L_\omega(\pi).
\label{eq:subproblem}
\end{equation}
The joint subproblem is solved with CP-SAT and iterative separation of worst-case probabilities.

Let $LB_k$ be a valid master lower bound and $UB_k$ the worst-case expected loss of the incumbent feasible policy $\pi^{\mathrm{best}}$. Then:
\begin{equation}
LB_k\le J_k^*\le UB_k
=\max_{p\in\mathcal P_k}\sum_\omega
p_\omega L_\omega(\pi^{\mathrm{best}}).
\label{eq:bounds}
\end{equation}

\subsubsection{Resource lower bounds and conditional cuts}
Conditioning on train--resource use extends single-resource scheduling lower bounds to different route combinations.

Let $\bar p_{i\omega}$ be the candidate route. Deviation from the candidate route combination is measured by:
\begin{equation}
\Delta_a(b,\bar a)=
\sum_{\omega\in\Omega_k}\sum_{i\in I_k^{\mathrm u}}
\left[1-a_{i,\bar p_{i\omega},\omega}(b)\right].
\label{eq:distance}
\end{equation}
A valid subproblem lower bound $\underline Q_k(\bar a)$ or a proof of route-combination infeasibility yields the following cuts, respectively:
\begin{equation}
\begin{aligned}
\theta&\ge\underline Q_k(\bar a)
[1-\Delta_a(b,\bar a)]
&&\text{(optimality cut)},\\
\Delta_a(b,\bar a)&\ge1
&&\text{(feasibility cut)}.
\end{aligned}
\label{eq:combinationcut}
\end{equation}

The train-use pattern of resource $r$ in the candidate route combination is:
\begin{equation}
\mathcal S_r(\bar a)=
\{(i,\omega):i\in I_k^{\mathrm u},\
r\in\mathcal R_{i\bar p_{i\omega}}\},
\label{eq:resource-support}
\end{equation}
Here, $\mathcal R_{ip}$ is the resource set of route $p$. In the single-resource relaxation, these trains may choose any feasible route using $r$. All other trains without fixed routes retain their full route sets:
\begin{equation}
P_{ir\omega}^{\mathrm{rel}}=
\begin{cases}
\{p_i^{\mathrm{fix}}\},&i\text{'s route is implemented or confirmed},\\
\{p\in P_i:r\in\mathcal R_{ip}\},&(i,\omega)\in\mathcal S_r(\bar a),\\
P_i,&\text{otherwise}.
\end{cases}
\label{eq:relaxed-routes}
\end{equation}
The relaxation retains ready times and required processing times for all trains. Inter-train mutual exclusion is retained only on resource $r$, and each scenario is solved separately. Let $\xi_{ip\omega}$ denote local route choices and mark other local variables with $\check{\cdot}$. The single-resource model is:
\begin{equation}
\begin{aligned}
Q_{r\omega}(\mathcal S_r)=
\min_{\xi,\check s,\check d,\check D}\quad&
\sum_{i\in I}\check D_{i\omega}\\
\text{s.t.}\quad&
\sum_{p\in P_{ir\omega}^{\mathrm{rel}}}\xi_{ip\omega}=1,
\qquad \xi_{ip\omega}\in\{0,1\},\\
&\check s_{i\omega}^{\mathrm{start}}\ge\max\{t_k,e_i\},
\qquad i\in I_k^{\mathrm u},\\
&\check d_{i\omega}\ge
\sum_{p\in P_{ir\omega}^{\mathrm{rel}}}w_{ip}\xi_{ip\omega},
\qquad \check d_{i\omega}=0\ \text{if }i\text{ is an originating train},\\
&\check C_{i\omega}=
\check s_{i\omega}^{\mathrm{start}}+
\sum_{p\in P_{ir\omega}^{\mathrm{rel}}}v_{ip}\xi_{ip\omega}
+\check d_{i\omega},\\
&\check D_{i\omega}\ge\check C_{i\omega}-C_i^0,
\qquad\check D_{i\omega}\ge0,\\
&\check{\mathcal I}_{ir\omega}\cap
\check{\mathcal I}_{jr\omega}=\varnothing,\qquad i\ne j,\\
&\check s_{ib\omega}\le a_r-1\ \text{or}\
\check s_{ib\omega}\ge b_{r\omega},
\qquad r(b)=r.
\end{aligned}
\label{eq:resource-relaxation}
\end{equation}
Local section times and reservation intervals follow Eqs.~\eqref{eq:block-link} and \eqref{eq:resource-intervals}. Section constraints apply only to selected routes. Actual train start times and section times already realized on resource $r$ are retained as in Eq.~\eqref{eq:executed}. Inter-train mutual exclusion on other resources, start-node time windows, and nonanticipativity across scenarios are relaxed.

The resource-use condition is expressed as:
\begin{equation}
\begin{aligned}
\ell_{i\omega}^{r}(b)&=
\sum_{n\in\operatorname{Path}(\omega)}
\sum_{p\in P_i:\,r\in\mathcal R_{ip}}x_{ipn},\\
\chi_r(b;\bar a)&=
\sum_{(i,\omega)\in\mathcal S_r(\bar a)}
\ell_{i\omega}^{r}(b)-|\mathcal S_r(\bar a)|+1.
\end{aligned}
\label{eq:resource-condition}
\end{equation}
The route-choice constraints imply $\ell_{i\omega}^{r}\in\{0,1\}$. If all resource-use relations in the candidate combination hold, $\chi_r=1$; otherwise, $\chi_r\le0$. Let $\underline L_{r\omega}\ge0$ be a valid lower bound obtained from Eq.~\eqref{eq:resource-relaxation}. We add the conditional cuts:
\begin{equation}
\begin{aligned}
\eta_\omega&\ge
\underline L_{r\omega}\chi_r(b;\bar a),
\qquad\omega\in\Omega_k,\\
\theta&\ge
\left[\max_{p\in\mathcal P_k}
\sum_{\omega\in\Omega_k}p_\omega\underline L_{r\omega}\right]
\chi_r(b;\bar a).
\end{aligned}
\label{eq:resourcecut}
\end{equation}

When $\chi_r=1$, any feasible schedule of the full problem satisfies the constraints retained in the single-resource model. Hence:
\begin{equation}
\underline L_{r\omega}
\le Q_{r\omega}(\mathcal S_r)
\le L_\omega(\pi),\qquad\omega\in\Omega_k.
\label{eq:resource-validity}
\end{equation}
Nonnegative probabilities allow maximization over $\mathcal P_k$ to yield a robust lower bound. When $\chi_r\le0$, the right-hand sides of the cuts are nonpositive. Their validity then follows from $\eta_\omega,\theta\ge0$.

\subsubsection{Graph learning-assisted resource ordering}
The lower-bound gain and solution time of a resource relaxation vary across resources. We construct a train--route--resource graph and learn the lower-bound gain per unit of solution time. Trains connect to their feasible routes, and routes connect to the resources they use.

Let $X$ be the node feature matrix. It encodes node types, ready and processing times, route choices, resource loads, and closure conditions. Let $A$ be the adjacency matrix with self-loops and $D_A$ its degree matrix. With weights $W_0,w_1$ and biases $b_0,b_1$, the network computes:
\begin{equation}
\begin{aligned}
\widetilde A&=D_A^{-1/2}AD_A^{-1/2},\\
H_G&=\tanh\!\left[(X+\widetilde AX)W_0+\boldsymbol1b_0^\top\right],\\
\widehat y&=(H_G+\widetilde AH_G)w_1+b_1\boldsymbol1.
\end{aligned}
\label{eq:graphscore}
\end{equation}
The output for resource node $r$ is denoted by $\widehat y_r$.

For training sample $\nu$, the resource relaxation yields a robust lower bound $\underline L_{\nu r}$ and solution time $c_{\nu r}$. Let $B_\nu$ be the nominal expected loss obtained from independent earliest train completions, and let $\epsilon_c$ be a lower threshold on solution time. The normalized supervision label is:
\begin{equation}
g_{\nu r}=
\frac{(\underline L_{\nu r}-B_\nu)^+}{\max\{\epsilon_c,c_{\nu r}\}},
\qquad
y_{\nu r}=\frac{g_{\nu r}}{\max\{1,\max_v g_{\nu v}\}}.
\label{eq:graph-label}
\end{equation}
For $N_{\mathrm{tr}}$ training samples with valid resource label sets $\mathcal R_\nu^{\mathrm{lab}}$, the training objective is:
\begin{equation}
\mathcal J=
\frac{1}{N_{\mathrm{tr}}}\sum_{\nu=1}^{N_{\mathrm{tr}}}
\frac{1}{|\mathcal R_\nu^{\mathrm{lab}}|}
\sum_{r\in\mathcal R_\nu^{\mathrm{lab}}}
(\widehat y_{\nu r}-y_{\nu r})^2.
\label{eq:graph-loss}
\end{equation}

During rescheduling, if $\rho>0$ and the incumbent feasible policy yields worst-case probabilities $p^*$, we calculate the resource risk load:
\begin{equation}
\lambda_r=
\sum_{\omega\in\Omega_k}p_\omega^*
\sum_{i\in I_k^{\mathrm u}}
\boldsymbol1\{r\in\mathcal R_{i\bar p_{i\omega}}\}.
\label{eq:risk-load}
\end{equation}
Resources are ordered by decreasing $\lambda_r$, with graph-predicted scores breaking ties. In other cases, resources are ordered by their graph-predicted scores.

\Needspace{4\baselineskip}
\par\noindent\textbf{Algorithm 1\quad GNN-LBBD solution algorithm}
\begin{enumerate}
\setlength{\itemsep}{0pt}
\renewcommand{\labelenumi}{\textbf{Step \arabic{enumi}:}}
\item Input the current information $\mathcal H_k$, actual operational state, reference schedule $P_0$, graph model, and computation budget. Construct conditional scenarios and information nodes. Generate an initial feasible policy using the scheduling heuristic. If the reopening time is known and the initial loss is positive, improve timings and sequences for the current routes. Compute $UB_k$ and set $LB_k=0$.
\item Build the master problem and retain previous cuts that remain valid for the current scenarios and operational state.
\item Solve the master problem and update $LB_k$. Separate worst-case probabilities using Eq.~\eqref{eq:probability-separation}. If a probability constraint is violated, add it to the master problem and return to Step 3.
\item Fix the candidate scenario routes and solve the joint scheduling subproblem. Update $UB_k$ using a feasible policy. Generate a route-combination cut from a valid lower bound or proven infeasibility.
\item Solve selected single-resource relaxations in resource-priority order within the remaining budget. Store scenario lower bounds and add resource conditional cuts.
\item Repeat Steps 3--5 until the computation budget is exhausted or $UB_k-LB_k\le\varepsilon$.
\item Return the incumbent best feasible policy and $LB_k,UB_k$.
\end{enumerate}

\section{Numerical experiments and results}
\subsection{Experimental design}
\label{sec:experiment-design}
\subsubsection{Test instances and settings}
The experiments use the public CP2025 in-station train dispatching benchmark of Schutt et al. The dataset contains 141 instances with 1--50 trains each. All instances use the same station with five platform tracks and 45 controlled sections. The inputs include train types, earliest ready times, feasible routes, section-occupancy sequences, and minimum running and dwell times.

The instances are divided into 85 training, 28 validation, and 28 test instances. Only the training instances are used for label generation and graph model training. The test set contains one instance for each train count from 1 to 22, and one each for 25, 30, 35, 40, 45, and 50 trains. The reference schedule $P_0$ is generated by a first-come, first-served rule. All algorithms start rescheduling from the same pre-disruption operational state.

Tabu search (TS) and variable neighborhood search (VNS) serve as the comparison algorithms. TS constructs neighborhoods through feasible rerouting, resource-order adjustments, and joint changes to small task subsets. It uses tabu memory and an aspiration criterion. VNS combines perturbation, local improvement, and neighborhood changes. Both algorithms use the same initial scheduling heuristic, permitted scheduling actions, and conditional predictions. Candidate solutions are evaluated with Eq.~\eqref{eq:objective}.

Table~\ref{tab:experiment-matrix} summarizes the experiment groups and comparisons. The representative instances cover fixed routes, alternative routes, and concentrated near-term operations. Each group considers platform-track and throat closures separately. Algorithm comparisons use the same computation budget for rescheduling.

\begin{table}[!htbp]\tableformat
\caption{Experimental groups and comparison settings}
\label{tab:experiment-matrix}
\begin{tabularx}{\linewidth}{@{}C{0.75}C{1}C{1.625}C{0.625}@{}}\toprule
Experiment & Instances & Comparison & Event seeds\\\midrule
Algorithm comparison & All 28 test instances & GNN-LBBD, TS, and VNS & 2001--2005\\
Disturbance combinations & t005-06, t006-06\newline t013-06, t019-06\newline t040-03, t045-03 & 8 presence/absence combinations of closures, arrival delays, and processing-time extensions & 2001--2005\\
Value of information & t005-06, t006-06\newline t013-06, t019-06\newline t040-03, t045-03 & With stage observations: scenario-based and conditional-mean predictions;\newline with closure-status observations only: scenario-based prediction & 2001--2005\\
Component ablation & t005-06, t019-06\newline t040-03, t045-03 & LBBD with resource lower bounds and cut strengthening, followed by GNN-assisted resource ordering & 2001--2005\\
Sensitivity analysis & t005-06, t019-06\newline t040-03, t045-03 & Closure-duration scale: $0.5$, $1$, $2$;\newline direct-increment scale: $0.5$, $1$, $1.5$ & 2001--2005\\
Distribution shift & t005-06, t019-06\newline t040-03, t045-03 & Actual closure-duration scale: $0.5$, $1$, $2$;\newline predicted scale: $1$; $\rho=0,\,0.2$ & 3001--3005\\
\bottomrule\end{tabularx}
\end{table}

\FloatBarrier
Let $C_i$ be the actual completion time of train $i$. Rescheduling outcomes are evaluated by total completion delay $L$, maximum train delay $D_{\max}$, on-time completion rate $P_{\mathrm{on}}$, final completion extension $E_{\mathrm{tail}}$, and delayed-train backlog $B(t)$:
\begin{equation}
\begin{aligned}
L&=\frac{1}{60}\sum_{i\in I}(C_i-C_i^0)^+,\qquad
D_{\max}=\frac{1}{60}\max_{i\in I}(C_i-C_i^0)^+,\\
P_{\mathrm{on}}&=\frac{|\{i\in I:C_i\le C_i^0\}|}{|I|},\qquad
E_{\mathrm{tail}}=\frac{(\max_i C_i-\max_i C_i^0)^+}{60},\\
B(t)&=\sum_{i\in I}\boldsymbol1\{C_i^0\le t<C_i\}.
\end{aligned}
\label{eq:backlog}
\end{equation}
The unit of $L$ is train-min. Both $D_{\max}$ and $E_{\mathrm{tail}}$ are measured in minutes, and $B(t)$ is measured in trains. A train is on time if it completes no later than in $P_0$. The external entry queue contains trains that are ready to enter but have not entered the station. Scheduling waiting time counts only waiting beyond required processing within the observation window.

Table~\ref{tab:algorithm-settings} lists the rescheduling and graph neural network training parameters.

\begin{table}[!htbp]\tableformat
\caption{Rescheduling and graph neural network training parameters}
\label{tab:algorithm-settings}
\begin{tabularx}{\linewidth}{@{}C{1.4}C{0.6}C{1.4}C{0.6}@{}}\toprule
Parameter & Value & Parameter & Value\\\midrule
Budget per rescheduling call (s) & 5 & Training scenarios & 510\\
Periodic check interval (s) & 120 & Supervision labels & 20880\\
Post-disruption observation window (s) & 7200 & Training epochs & 200\\
Maximum rescheduling rounds & 160 & Learning rate & 0.001\\
Prediction and search budget share (\%) & 95 & Algorithm seeds & 101, 211\\
Conditional stratified scenarios & 6 & Probability radius $\rho$ & 0.2\\
Input feature dimension & 8 & Hidden dimension & 12\\
Resources checked per round & 2 & Label time threshold (s) & 0.01\\
Single-resource time limit (s) & 1 & Termination tolerance (s) & $10^{-6}$\\
\bottomrule\end{tabularx}
\end{table}

The experiments were run on Windows 11 with an Intel64 Family 6 Model 183 processor (32 logical processors), CPython 3.12.14, and OR-Tools 9.15.6755. Each run used one thread, with at most four runs in parallel. Inference during rescheduling was included in the computation budget.

\subsubsection{Disturbance design}
\label{sec:disturbance-design}
We introduce resource closures $R$, arrival delays $A$, and required processing-time extensions $D$ into $P_0$ to examine resource restrictions and changes in train operating conditions jointly. Each event closes one platform track or one throat resource. The designed number of affected trains ranges from 20\% to 60\% of the eligible trains for arrival delays and processing-time extensions. The corresponding per-train increments range from 0 to 300~s and from 0 to 200~s. Nominal closure-duration scales are 300, 600, and 1200~s for low, medium, and high disturbance intensities, respectively.

A Bayesian network links the three disturbances through a common latent intensity $Z\in\{\mathrm L,\mathrm M,\mathrm H\}$ as the parent node. Let $\mathcal D_A$ and $\mathcal D_D$ contain the affected train sets and their direct increments, and let $T_R$ be the closure duration. Their joint distribution is:
\begin{equation}
\Pr(\mathcal D_A,\mathcal D_D,T_R)
=\sum_z\pi_z\Pr(\mathcal D_A\mid z)\Pr(\mathcal D_D\mid z)\Pr(T_R\mid z),
\label{eq:joint-disturbance}
\end{equation}
Here, $\pi_z=\Pr(Z=z)$.

The event time $t_0$ is selected using the number of planned operation starts in the following 600~s. Let $s_i^0$ and $p_i^0$ be the planned start time and route of train $i$. The eligible sets for arrival delays and processing-time extensions are:
\begin{equation}
\begin{aligned}
\mathcal E_A(t_0)&=\{i\in I:\ i\text{ is an external non-originating train},\ s_i^0\ge t_0\},\\
\mathcal E_D(t_0)&=\{i\in\mathcal E_A(t_0):\underline d_{i p_i^0}>0\}.
\end{aligned}
\label{eq:eligible-pools}
\end{equation}

The closed resource is selected using its planned usage in the 600~s after the event. A platform-track closure covers its three platform sections. Conditional on $Z=z$, the affected sets $\mathcal U_A$ and $\mathcal U_D$ are sampled uniformly without replacement from $\mathcal E_A$ and $\mathcal E_D$, respectively. The designed numbers of affected trains and their increments are:
\begin{equation}
\begin{aligned}
n_X(z)&=\lceil\alpha_z|\mathcal E_X|\rceil,\qquad X\in\{A,D\},\\
\widetilde\Delta_i^A\mid Z=z&\sim\operatorname{TN}_{[0,300]}\!\left(150s_z,(60s_z)^2\right),\quad i\in\mathcal U_A,\\
\widetilde\Delta_i^D\mid Z=z&\sim\operatorname{TN}_{[0,200]}\!\left(100s_z,(40s_z)^2\right),\quad i\in\mathcal U_D.
\end{aligned}
\label{eq:increment-generation}
\end{equation}
The notation $\operatorname{TN}_{[l,u]}(\mu,\sigma^2)$ denotes a truncated normal distribution, where $\mu$ and $\sigma$ are the mean and standard deviation before truncation. Increments are sampled independently across trains and rounded to seconds. Table~\ref{tab:settings} lists the intensity parameters.

\begin{table}[H]\tableformat
\caption{Disturbance intensity and distribution parameters}
\label{tab:settings}
\begin{tabularx}{\linewidth}{@{}C{0.75}C{0.875}C{1}C{1.375}C{1}@{}}\toprule
Intensity $z$ & Probability $\pi_z$ & Increment factor $s_z$ & Closure-duration scale $m_z$ (s) & Affected proportion $\alpha_z$\\\midrule
Low & 0.25 & 0.5 & 300 & 0.20\\
Medium & 0.50 & 1.0 & 600 & 0.40\\
High & 0.25 & 2.0 & 1200 & 0.60\\
\bottomrule\end{tabularx}
\end{table}

Disruption resolution is represented by an absorbing Markov chain. The state $S(t)\in\{1,2,3,4\}$ comprises three sequential transient states and an absorbing state for resource reopening. The initial state is 1. With a time step of 1~s and intensity $z$, the process satisfies:
\begin{equation}
\begin{aligned}
P_z&=\begin{pmatrix}
1-q_z&q_z&0&0\\
0&1-q_z&q_z&0\\
0&0&1-q_z&q_z\\
0&0&0&1
\end{pmatrix},\qquad q_z=1-\exp(-3/m_z),\\
T_R&=\sum_{j=1}^{3}G_{jz},\qquad
G_{jz}\overset{\mathrm{iid}}{\sim}\operatorname{Geom}(q_z).
\end{aligned}
\label{eq:recovery}
\end{equation}
The nominal closure-duration scale $m_z$ is given in Table~\ref{tab:settings}. Since $G_{jz}$ takes positive integer values, $\mathbb E[T_R\mid z]=3/q_z$.

Stage observations identify the disruption-resolution stage at the collection time. Matrix $H$ describes their misclassification. Observation $k$ is collected at $\tau_k$ and received at $a_k$ after a delay $L_k$. Observations are collected every 120~s. The delay follows a discrete uniform distribution over 0--60~s:
\begin{equation}
\begin{aligned}
\tau_k&=t_0+120k,\quad a_k=\tau_k+L_k,\quad k\ge1,\quad \tau_k<t_0+T_R,\\
L_k&\sim\operatorname{DU}\{0,\ldots,60\}.
\end{aligned}
\label{eq:report-times}
\end{equation}
Let $H_{jv}$ be the probability of observing stage $v$ when the true stage is $j$. The stage misclassification matrix is:
\begin{equation}
H=\begin{pmatrix}0.8&0.1&0.1\\0.1&0.8&0.1\\0.1&0.1&0.8\end{pmatrix}.
\label{eq:report-noise}
\end{equation}

Initial arrival delays and processing-time increments provide evidence about disturbance intensity. Stage observations further update the predicted remaining closure duration. Let $\mathcal D_0$ be the initial evidence and $\mathcal L_z(\mathcal D_0)$ its likelihood under intensity $z$.

\Needspace{6\baselineskip}
Let $Y_\ell$ be stage observation $\ell$. The observations available by $t_k$ are \mbox{$\mathcal O_k=\{(\tau_\ell,Y_\ell):a_\ell\le t_k\}$}. While the resource remains closed, the joint posterior of intensity and current stage is:
\begin{equation}
\begin{aligned}
\varphi_k(z,j)&=\pi_z\mathcal L_z(\mathcal D_0)
\Pr(S(t_k)=j,\mathcal O_k\mid Z=z),\\
\beta_k(z,j)&=\frac{\varphi_k(z,j)}
{\sum_{z'}\sum_{h=1}^{3}\varphi_k(z',h)},\qquad j=1,2,3.
\end{aligned}
\label{eq:belief}
\end{equation}
Bayesian filtering combines $P_z$ and $H$ to update the state distribution at observation collection times and propagate it to the current decision time. The initial evidence likelihood is computed from the number of affected trains and the rounded increments.

Let $Q_z$ be the transient submatrix of $P_z$, $\boldsymbol\beta_k(z)=(\beta_k(z,1),\beta_k(z,2),\beta_k(z,3))$, and $I_3$ the identity matrix of order three. The conditional distribution and mean of the remaining closure duration satisfy:
\begin{equation}
\begin{aligned}
\Pr(R_k>u\mid\mathcal H_k)
&=\sum_z\boldsymbol\beta_k(z)Q_z^u\boldsymbol1,
\quad u\in\mathbb Z_{\ge0},\\
\mathbb E[R_k\mid\mathcal H_k]
&=\sum_z\boldsymbol\beta_k(z)(I_3-Q_z)^{-1}\boldsymbol1.
\end{aligned}
\label{eq:remaining-time}
\end{equation}
Scenario-based prediction uses stratified sampling over conditional quantile intervals to generate reopening times and subsequent observation sequences. Each scenario has nominal probability $1/|\Omega_k|$. Conditional-mean prediction represents the reopening time using the mean in Eq.~\eqref{eq:remaining-time}.
\subsection{Algorithm comparison and ablation analysis}
\label{sec:overall-results}
\Needspace{16\baselineskip}
\subsubsection{Rescheduling performance and computation time}
\begin{figure}[!htbp]\centering
\includegraphics[scale=0.75]{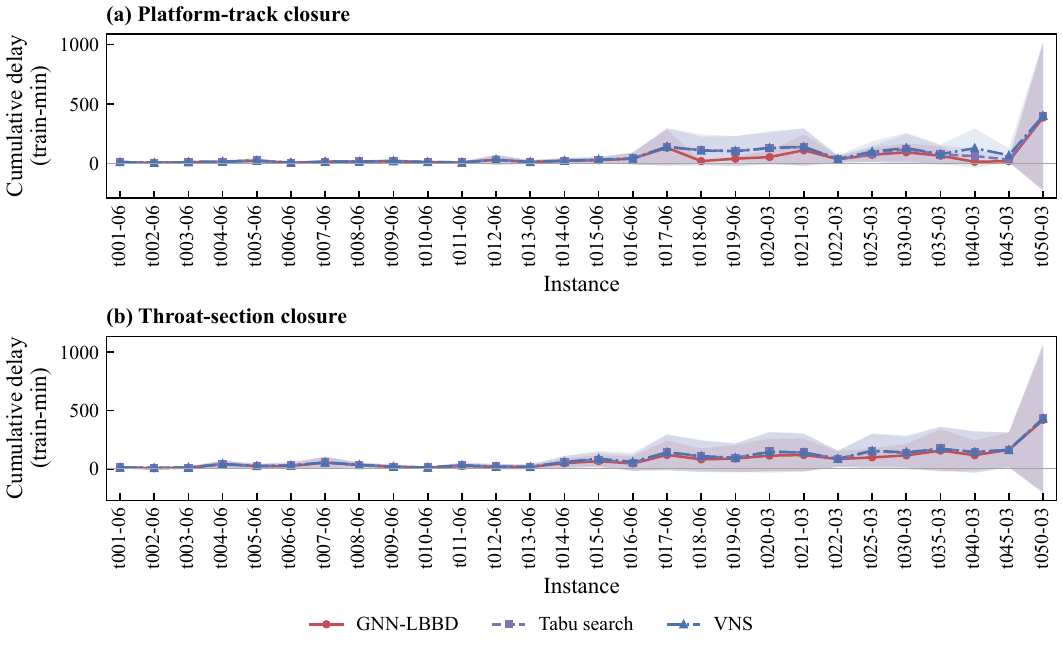}
\caption{Total completion delay by algorithm}
\label{fig:delay}
\end{figure}

\begin{figure}[H]\centering
\includegraphics[scale=0.75]{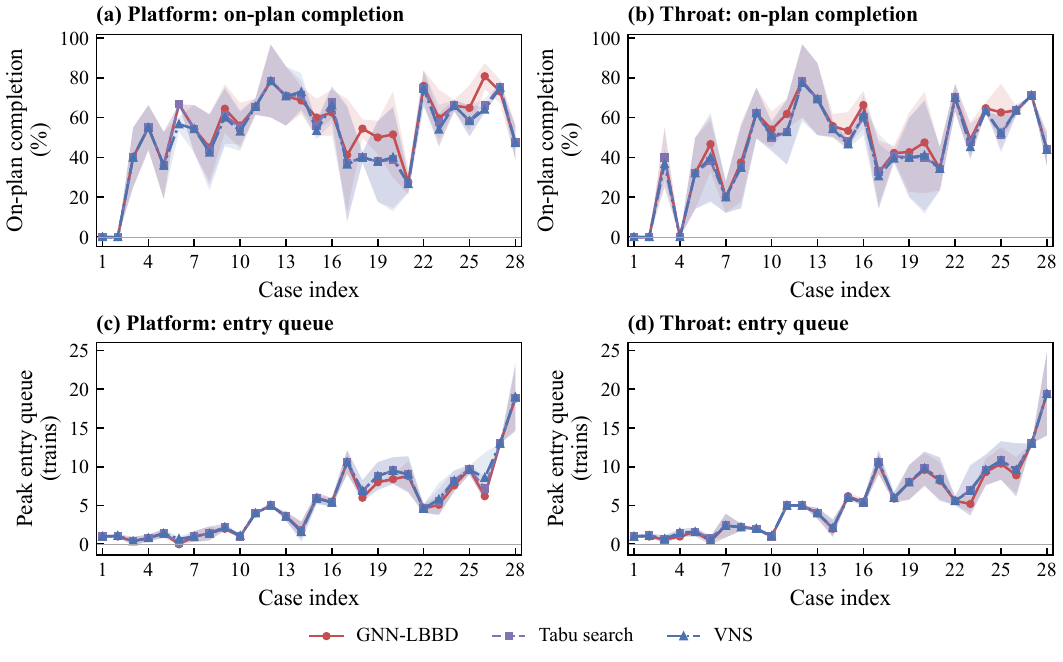}
\caption{On-time completion rates and maximum external entry queues}
\label{fig:operations}
\end{figure}

\begin{figure}[H]\centering
\includegraphics[scale=0.75]{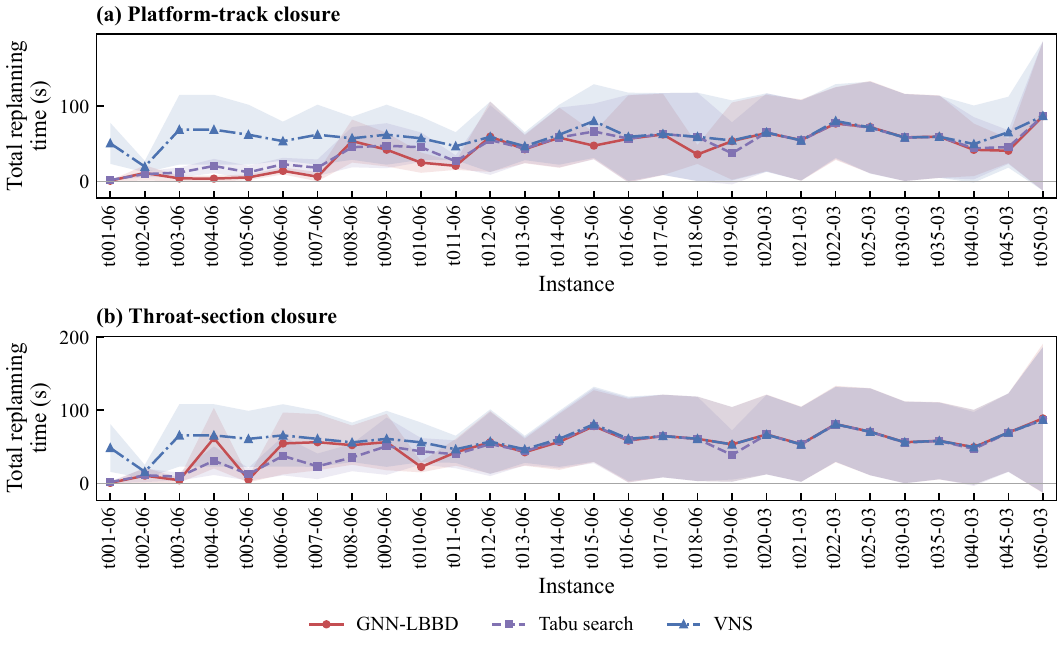}
\caption{Total rescheduling computation time by algorithm}
\label{fig:runtime}
\end{figure}

\begin{figure}[H]\centering
\includegraphics[width=0.84\linewidth,trim=0bp 20bp 623.622bp 0bp,clip]{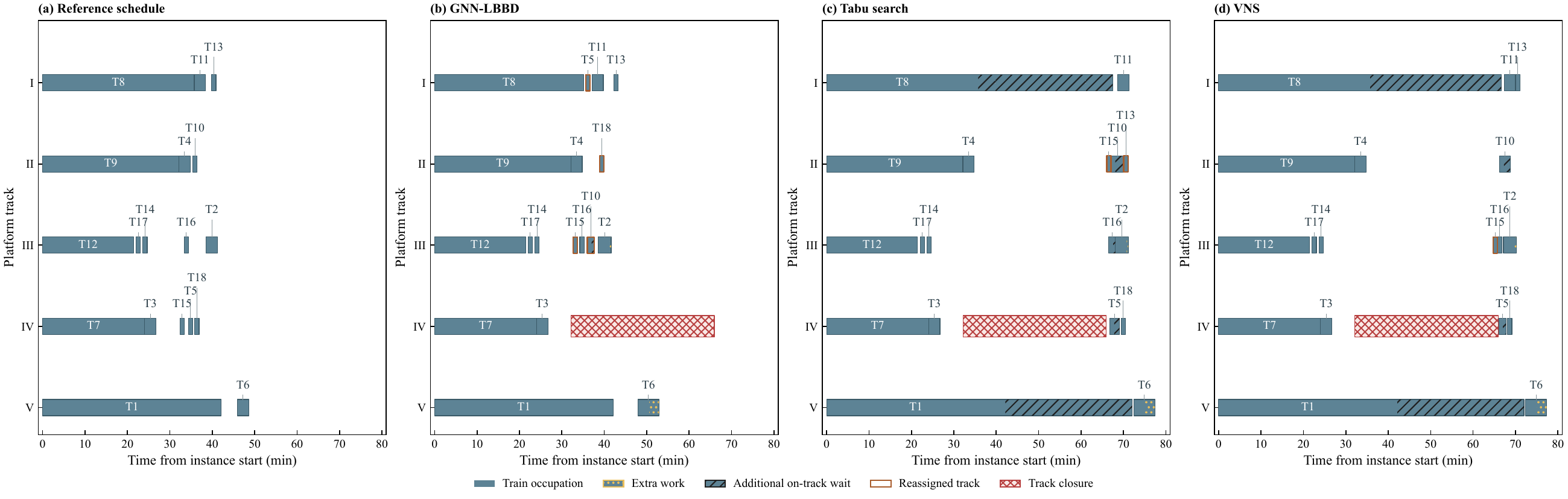}\\[3pt]
\includegraphics[width=0.84\linewidth,trim=623.622bp 20bp 0bp 0bp,clip]{F12_platform_occupation_case.pdf}\\[2pt]
\includegraphics[width=0.82\linewidth,trim=362bp 4bp 372.244bp 376bp,clip]{F12_platform_occupation_case.pdf}
\caption{Platform-track occupation under different algorithms}
\label{fig:occupation}
\end{figure}

We compare GNN-LBBD, TS, and VNS under identical reference schedules, disturbances, and computation budgets to assess their delay-reduction performance. Table~\ref{tab:main} and Figs.~\ref{fig:delay}--\ref{fig:runtime} summarize scheduling and computational results. Figure~\ref{fig:occupation} compares platform-track occupation under the same closure event.

\begin{table}[H]\tableformat
\caption{Rescheduling performance of the algorithms}
\label{tab:main}
\begin{tabularx}{\linewidth}{@{}C{1.84}*{3}{C{0.72}}@{}}\toprule
Metric & GNN-LBBD & TS & VNS\\\midrule
Total completion delay (train-min) & 61.13 & 73.33 & 76.47\\
Maximum train delay (min) & 10.31 & 10.53 & 10.70\\
Delayed trains (trains) & 7.30 & 7.79 & 7.86\\
On-time completion rate (\%) & 50.30 & 47.96 & 47.46\\
Final completion extension (min) & 5.79 & 6.42 & 6.65\\
Total scheduling waiting time (train-min) & 83.47 & 95.83 & 98.97\\
Maximum delayed-train backlog (trains) & 4.25 & 4.82 & 4.93\\
Mean entry delay (min) & 3.63 & 4.21 & 4.31\\
Total rescheduling computation time (s) & 46.15 & 46.42 & 59.91\\
\bottomrule\end{tabularx}
\end{table}

GNN-LBBD primarily reduced total completion delay and scheduling waiting time. In Table~\ref{tab:main}, its mean total completion delay was 61.13 train-min, 16.63\% below TS and 20.06\% below VNS. Its total scheduling waiting time was 83.47 train-min, compared with 95.83 and 98.97 train-min for the two comparison algorithms. The on-time completion rate increased to 50.30\%, from 47.96\% for TS and 47.46\% for VNS. The final completion extension decreased to 5.79~min, from 6.42 and 6.65~min, respectively.

GNN-LBBD improved schedule quality with a total computation time close to that of TS. In Table~\ref{tab:main}, mean total computation times were 46.15~s for GNN-LBBD, 46.42~s for TS, and 59.91~s for VNS. At similar total computation times, GNN-LBBD reduced total completion delay by 16.63\% relative to TS. This indicates better schedule quality within a limited computation budget.

\FloatBarrier

Track reassignment and coordination between successive trains allowed some operations to finish before reopening. Consider the platform-track closure in t018-06 (Fig.~\ref{fig:occupation}). GNN-LBBD reassigned T15, T5, and T18 from the closed track IV to tracks III, I, and II, respectively. It also moved T10 from track II to track III. On track I, T8 departed before T5 passed through and T11 entered. On track V, T6 followed T1. Under TS and VNS, T8 and T1 each incurred about 30~min of additional on-track waiting. T11 and T6 entered their tracks only after reopening. GNN-LBBD completed these operations during the closure by coordinating resource release, rerouted train movements, and subsequent arrivals. The benefits extended to T11 and T6, which did not originally use the closed track. This reduced the propagation of closure-induced delays through successive track occupations.
\FloatBarrier

\Needspace{5\baselineskip}
\subsubsection{Component ablation experiments}
\label{sec:components}
\begin{figure}[!htbp]\centering
\includegraphics[scale=0.75]{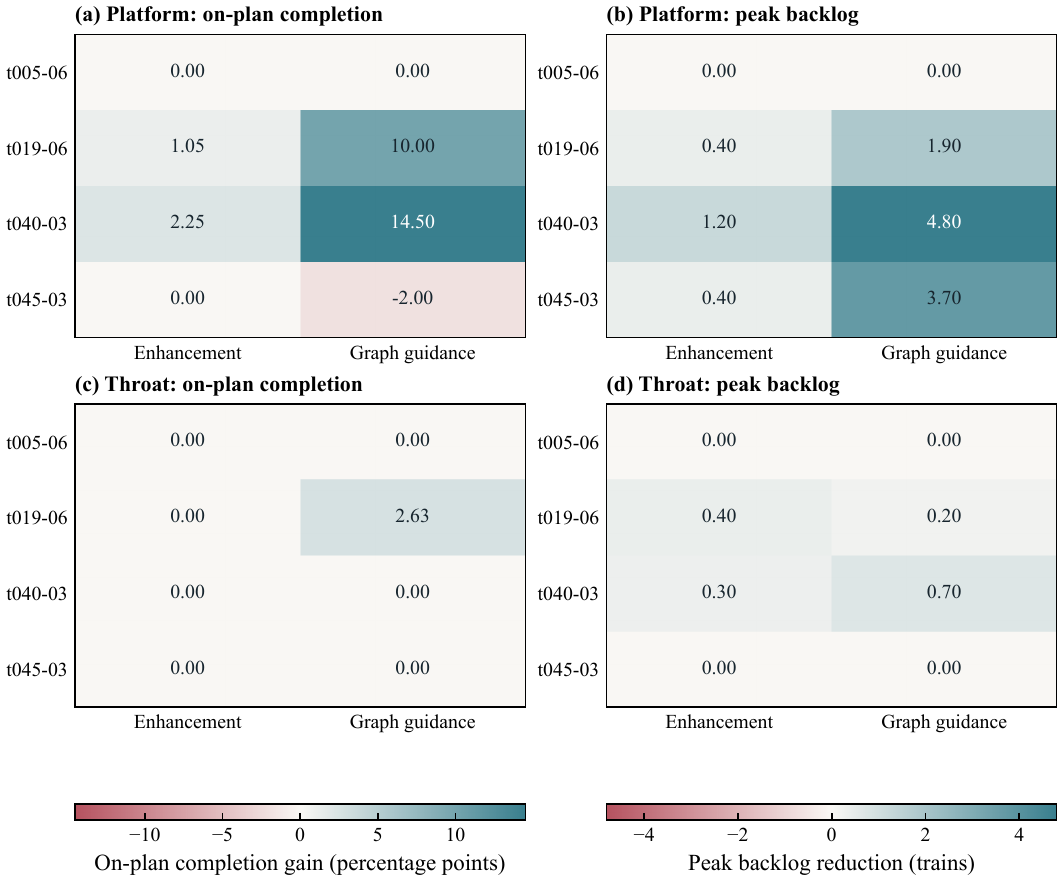}
\caption{Effects of algorithm components on on-time completion and maximum backlog}
\label{fig:ablation}
\end{figure}

We compare configurations that progressively add resource lower bounds with cut strengthening and graph learning-assisted resource ordering. This isolates their contributions within a limited computation budget. Table~\ref{tab:ablation} summarizes scheduling metrics and paired event comparisons. Figure~\ref{fig:ablation} shows changes in the on-time completion rate and maximum backlog.

\begin{table}[!htbp]\tableformat
\caption{LBBD component ablation results}
\label{tab:ablation}
\begin{tabularx}{\linewidth}{@{}C{0.95}C{0.95}C{1.05}C{0.85}C{0.65}C{1.55}@{}}\toprule
Resource lower bounds\newline and cut strengthening & GNN-assisted\newline resource ordering & Total completion\newline delay (train-min) & Maximum train\newline delay (min) & Maximum backlog\newline (trains) & Lower/unchanged/higher delay\newline vs. previous configuration\newline (events)\\\midrule
Disabled & Disabled & 98.88 & 12.47 & 6.58 & ---\\
Enabled & Disabled & 85.15 & 12.16 & 6.24 & 19/19/2\\
Enabled & Enabled & 60.90 & 9.98 & 4.83 & 20/20/0\\
\bottomrule\end{tabularx}
\end{table}

Resource lower bounds with cut strengthening reduced total completion delay, and graph learning-assisted ordering provided a further improvement. In Table~\ref{tab:ablation}, enabling resource lower bounds and cut strengthening reduced mean total completion delay from 98.88 to 85.15 train-min, or 13.88\%. Adding graph learning-assisted ordering further reduced it to 60.90 train-min, 28.48\% below the configuration without graph learning. Across 40 paired events, resource lower bounds with cut strengthening reduced delay in 19 events. Adding graph learning-assisted ordering reduced delay in 20 events. Its incremental benefit was therefore supported across multiple events.

Graph learning-assisted ordering also reduced peak backlog and individual train delay. In Table~\ref{tab:ablation}, the mean maximum backlog decreased from 6.24 to 4.83 trains, and the mean maximum train delay decreased from 12.16 to 9.98~min. Under platform-track closures, maximum backlogs in the 19- and 40-train instances decreased by 1.90 and 4.80 trains (Fig.~\ref{fig:ablation}). Graph learning-assisted ordering further reduced total completion delay after resource lower bounds and cut strengthening had been enabled. The solution order of resource relaxations is thus an effective component for improving schedule quality within a limited computation budget.

\Needspace{5\baselineskip}
\FloatBarrier
\subsection{Combined disturbances and information updates}
\subsubsection{Effects of disturbance combinations}
We compare eight combinations of three disturbances to examine whether resource closures change the delay effects of arrival delays and processing-time extensions. Table~\ref{tab:combinations} summarizes mean total completion delay. Figure~\ref{fig:combinations} shows delay distributions across instances and closure locations.

\begin{table}[H]\tableformat
\caption{Total completion delay under disturbance combinations}
\label{tab:combinations}
\begin{tabularx}{\linewidth}{@{}C{1.2}C{0.8}C{1.2}C{0.8}@{}}\toprule
Disturbance combination & Total completion delay (train-min) & Disturbance combination & Total completion delay (train-min)\\\midrule
No disturbance & 0.00 & Closure and arrival delays & 57.64\\
Resource closure & 54.46 & Closure and processing-time extensions & 60.20\\
Arrival delays & 4.04 & Arrival delays and processing-time extensions & 7.76\\
Processing-time extensions & 3.83 & All three disturbances & 45.41\\
\bottomrule\end{tabularx}
\end{table}

Resource closures changed the relative effects of arrival delays and processing-time extensions on total completion delay. Without closures, arrival delays and processing-time extensions alone produced mean total completion delays of 4.04 and 3.83 train-min, respectively. Their combined loss was 7.76 train-min, close to the sum of the individual losses. Under closures, mean total completion delay increased from 54.46 train-min to 57.64 with arrival delays and 60.20 with processing-time extensions. The two disturbances caused similar losses without closures, but processing-time extensions produced a larger mean additional delay when combined with closures. Their relative effects under closures should therefore not be inferred directly from their individual losses without closures.

Mean total completion delay under all three disturbances was lower than that under closure alone. The combined loss was 45.41 train-min, also below those for closures combined separately with arrival delays or processing-time extensions. Without closures, the joint loss of the two time disturbances was close to the sum of their individual losses. Under closures, adding either disturbance separately increased mean delay, whereas adding both reduced it. The combined losses therefore departed from an approximately additive relationship.

\edef\landscapepagenumberx{\the\dimexpr\paperwidth-\paperheight+1in+\topmargin+\headheight+\headsep+\textheight+\footskip\relax}
\begin{landscape}
\thispagestyle{empty}
\AddToHookNext{shipout/foreground}{%
\put(\landscapepagenumberx,-.5\paperheight){\rotatebox{90}{\makebox[0pt][c]{\normalfont\thepage}}}%
}
\begin{figure}[p]\centering
\includegraphics[width=\linewidth]{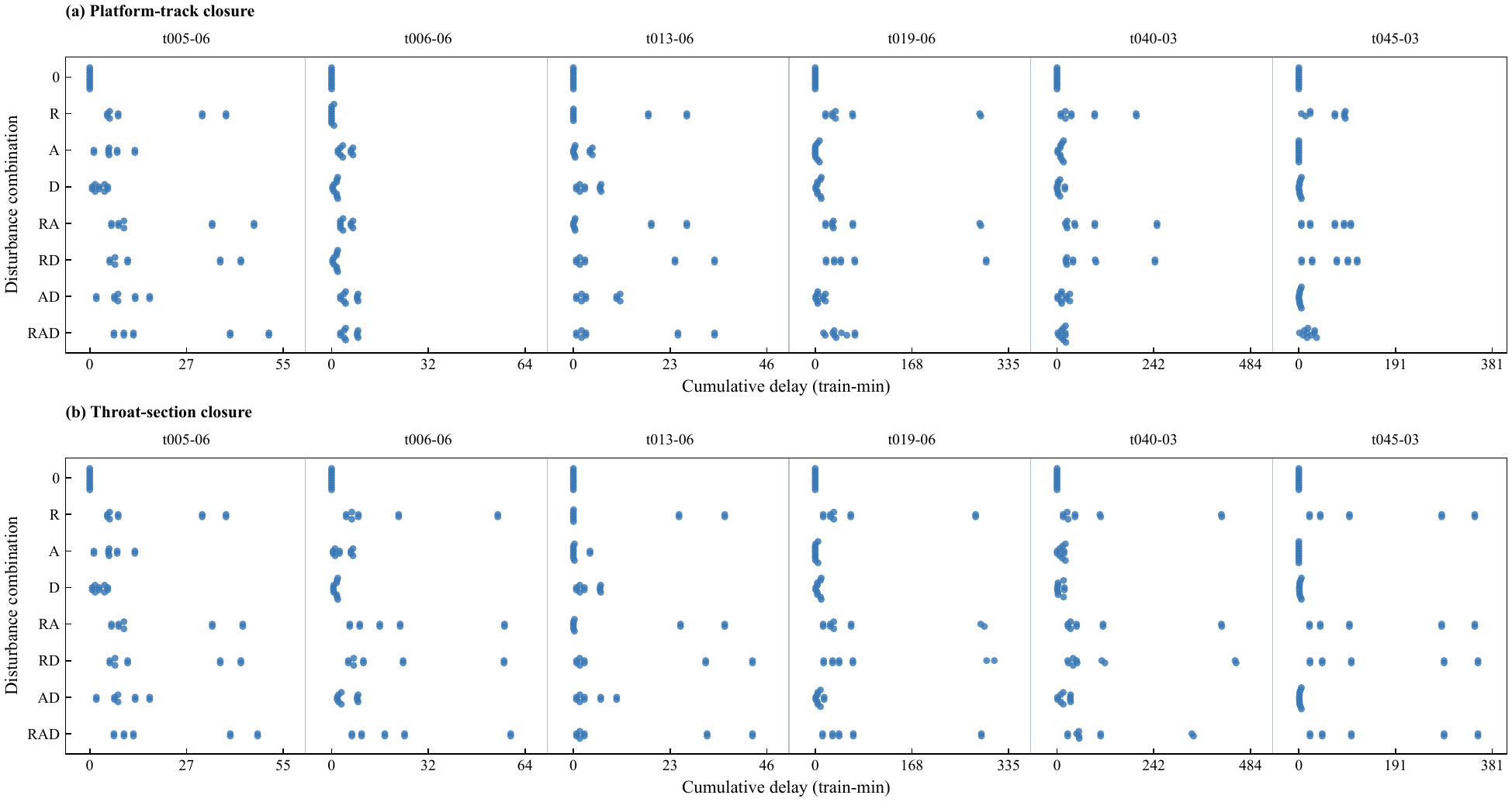}
\caption{Total completion delay distributions under disturbance combinations}
\label{fig:combinations}
\end{figure}
\end{landscape}

\subsubsection{Effects of reopening-time prediction on rescheduling}
\begin{figure}[!htbp]\centering
\includegraphics[scale=0.75]{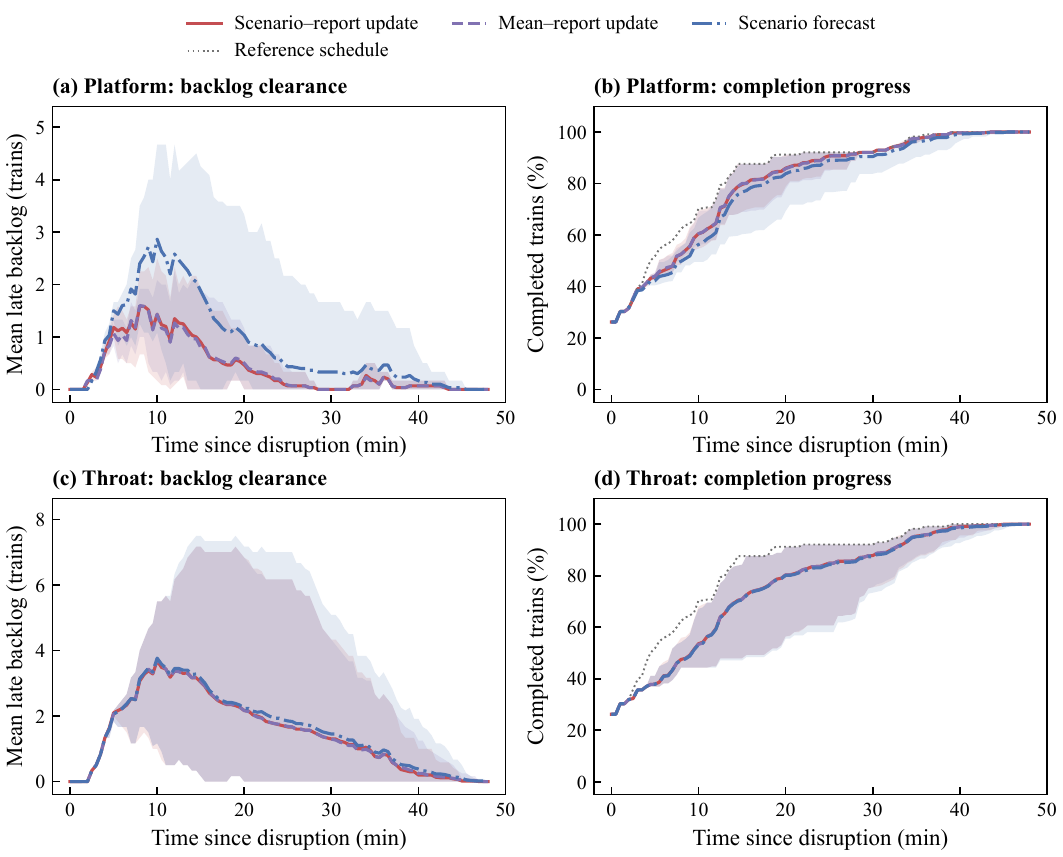}
\caption{Train backlog and completion progress under different prediction schemes}
\label{fig:recoveryprogress}
\end{figure}

We compare predictions with and without stage observations, and scenario-based with conditional-mean predictions under identical observations. These comparisons assess how observations and prediction representations affect train holding and track assignments. Table~\ref{tab:information} summarizes total completion delay, and Fig.~\ref{fig:recoveryprogress} shows train backlog and completion progress. Execution records from the same closure event are used to examine scheduling differences.

\begin{table}[!htbp]\tableformat
\caption{Total completion delay under different prediction schemes}
\label{tab:information}
\begin{tabularx}{\linewidth}{@{}C{0.875}C{1}C{1.125}C{1}@{}}\toprule
\multirow{2}{*}{Instance} & \multicolumn{3}{c}{Total completion delay (train-min)}\\
\cmidrule(l){2-4}
 & Scenario-based\newline with stage observations & Conditional mean\newline with stage observations & Scenario-based\newline with closure-status observations only\\\midrule
t005-06 & 23.56 & 23.56 & 23.56\\
t006-06 & 14.66 & 14.66 & 14.66\\
t013-06 & 14.22 & 14.22 & 14.21\\
t019-06 & 63.93 & 64.04 & 91.66\\
t040-03 & 65.41 & 66.75 & 102.99\\
t045-03 & 90.71 & 89.05 & 100.77\\\midrule
Mean & 45.41 & 45.38 & 57.97\\
\bottomrule\end{tabularx}
\end{table}

Rescheduling with stage observations achieved lower total completion delay. In Table~\ref{tab:information}, scenario-based prediction with stage observations reduced mean total completion delay from 57.97 to 45.41 train-min, or approximately 21.7\%. The comparison scheme observed only the closure status. About 10~min after a platform-track closure, its mean backlog approached three trains, compared with approximately 1.5 for both schemes using stage observations (Fig.~\ref{fig:recoveryprogress}). The latter schemes also achieved higher completion rates during the same period.

Prediction differences also changed track assignments during the closure. Consider the throat-closure event in t040-03, with time measured from closure onset. At 2.08~min, an observation indicated that the first resolution stage was not yet complete. The estimated reopening time changed from 20.09 to 21.71~min after closure onset. Later observations continued to revise this estimate. At 8.15~min, the predicted remaining closure duration was 13.23~min. Train 25 was subsequently reassigned from track IV to track V. It entered at 9.15~min and completed its operations at 11.15~min. With closure-status observations only, the train retained track IV and entered at 34.58~min, incurring 25.43~min of additional scheduling waiting. Both solver runs produced these track-assignment and entry-time differences. The resource reopened at 29.25~min. The scheme using stage observations had already completed the train's operations, whereas the other scheme continued waiting until after reopening.

With identical stage observations, scenario-based and conditional-mean predictions yielded mean total completion delays of 45.41 and 45.38 train-min, respectively. The delay difference associated with using stage observations was larger than the mean difference between the two prediction representations. In terms of mean operational loss in these experiments, continued use of stage observations deserves priority over refining the prediction representation.

\subsection{Sensitivity and distribution-shift analyses}
\subsubsection{Sensitivity to disturbance scales}
\begin{figure}[!htbp]\centering
\includegraphics[width=\linewidth]{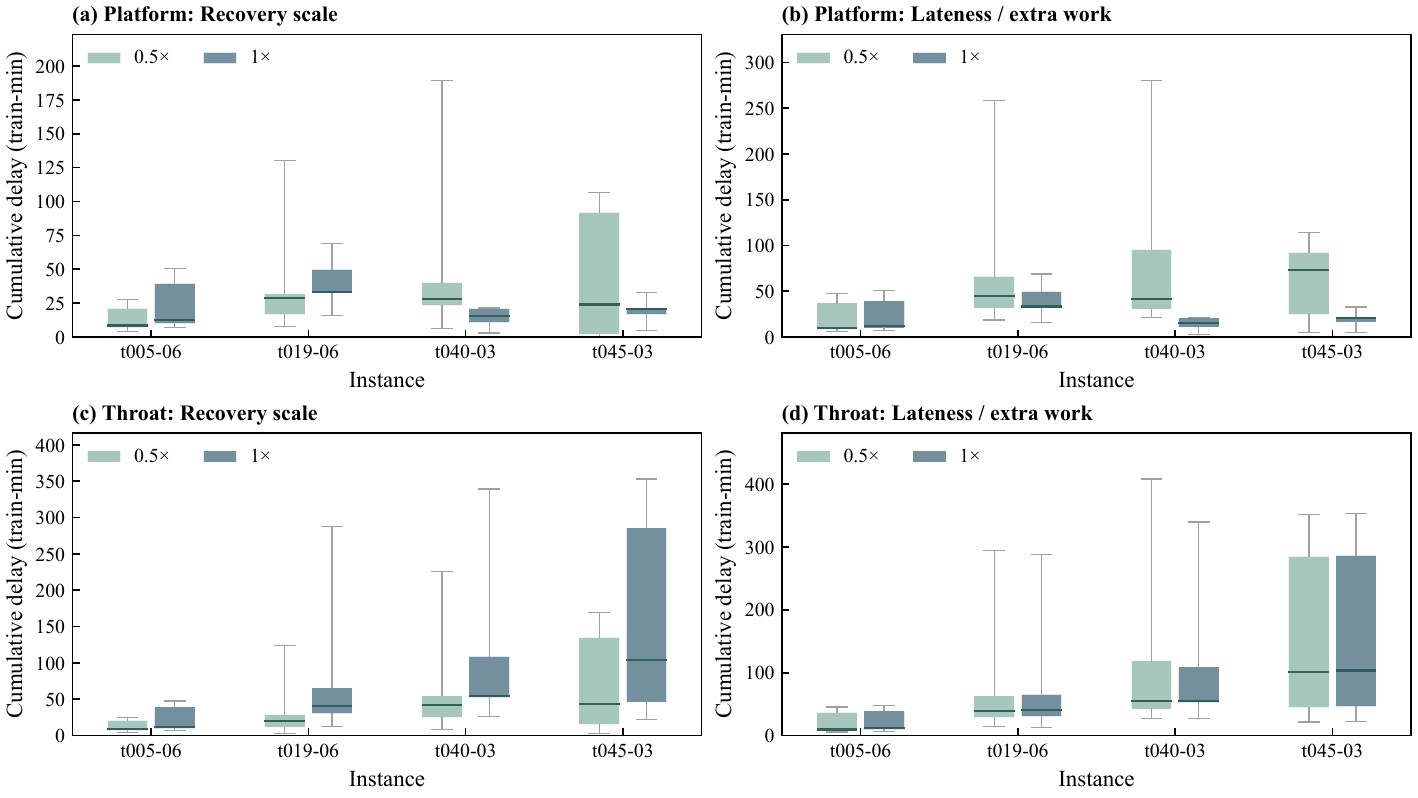}
\caption{Total completion delay distributions under different disturbance scales}
\label{fig:sensitivity}
\end{figure}

We vary the closure-duration scale and the arrival-delay and processing-time increment scale separately to examine how disturbance scales relate to rescheduling losses. The same parameters are used for actual disturbance generation and prediction. Table~\ref{tab:sensitivity} summarizes rescheduling metrics, and Fig.~\ref{fig:sensitivity} compares total completion delay distributions at half and baseline scales.

\begin{table}[H]\tableformat
\caption{Rescheduling performance under different disturbance scales}
\label{tab:sensitivity}
\begin{tabularx}{\linewidth}{@{}C{1.5}C{1.0}C{0.85}C{0.65}@{}}\toprule
Setting & Total completion delay (train-min) & Maximum train delay (min) & Maximum backlog (trains)\\\midrule
Baseline & 60.90 & 9.98 & 4.83\\
Closure-duration scale $0.5$ & 44.26 & 7.02 & 4.53\\
Closure-duration scale $2$ & 170.60 & 21.43 & 8.21\\
Arrival-delay and processing-time increments $0.5$ & 82.60 & 11.84 & 6.20\\
Arrival-delay and processing-time increments $1.5$ & 83.36 & 12.17 & 6.04\\
\bottomrule\end{tabularx}
\end{table}

Longer closures were associated with a more than proportional increase in total completion delay. In Table~\ref{tab:sensitivity}, doubling the closure-duration scale increased mean total completion delay from 60.90 to 170.60 train-min, or 2.80 times the baseline. Maximum train delay increased from 9.98 to 21.43~min, and maximum backlog increased from 4.83 to 8.21 trains. Halving the duration scale reduced mean total completion delay to 44.26 train-min, 27.3\% below the baseline. Longer closures increased both backlog size and individual train delay. After reopening, operation sequences remain constrained by shared-resource occupancy. Operations postponed during a closure may therefore continue to generate queues. Proportional extrapolation from baseline delay may underestimate the cumulative effects of long closures on subsequent operations.

Smaller direct time increments did not produce a corresponding reduction in total delay. At arrival-delay and processing-time increment scales of $0.5$ and $1.5$, mean total completion delays were 82.60 and 83.36 train-min, respectively. Both exceeded the baseline of 60.90 train-min. Direct increments change train ready times and required processing times. Rescheduling losses also depend on route choices and coordination between operations. Disturbance magnitude and actual losses therefore did not follow a simple proportional relationship under the given rescheduling strategy. Estimating operating benefits from the reduction in input increments may overstate the corresponding delay improvement.

\subsubsection{Closure-duration distribution shifts}
\begin{figure}[!htbp]\centering
\includegraphics[width=\linewidth]{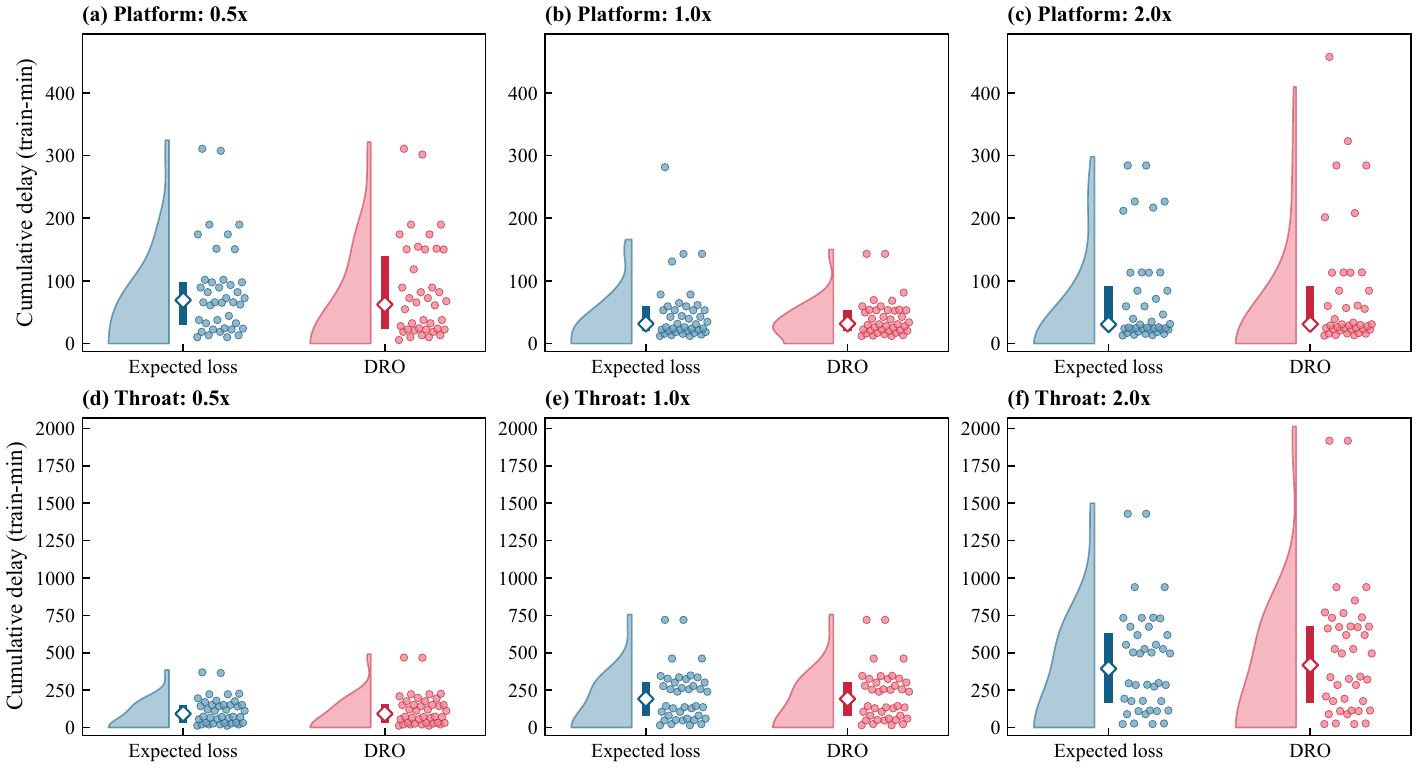}
\caption{Total completion delay distributions under closure-duration distribution shifts}
\label{fig:robustness}
\end{figure}

We compare the expected-loss and distributionally robust models when actual closure durations deviate from predictions. The predicted duration scale remains at its baseline value, while the actual scale is set to $0.5$, $1$, or $2$. Table~\ref{tab:robustness} summarizes rescheduling metrics. Figure~\ref{fig:robustness} shows total completion delay distributions across closure locations and actual duration scales.

\begin{table}[H]\tableformat
\caption{Model performance under closure-duration distribution shifts}
\label{tab:robustness}
\begin{tabularx}{\linewidth}{@{}C{0.6}C{1.15}C{1.25}C{1.2}C{0.8}@{}}\toprule
Actual scale & Model & Total completion delay (train-min) & Maximum train delay (min) & Maximum backlog (trains)\\\midrule
\multirow{2}{*}{$0.5$} & Expected-loss & 97.64 & 13.97 & 7.56\\
 & Distributionally robust & 100.92 & 13.92 & 7.41\\
\multirow{2}{*}{$1$} & Expected-loss & 132.09 & 19.69 & 6.76\\
 & Distributionally robust & 127.98 & 19.28 & 6.63\\
\multirow{2}{*}{$2$} & Expected-loss & 255.42 & 36.88 & 6.86\\
 & Distributionally robust & 284.37 & 36.88 & 7.46\\
\bottomrule\end{tabularx}
\end{table}

Distributionally robust evaluation changes the trade-off between mean predicted delay and losses under probability deviations. Consider the throat-closure event in t040-03 at the same decision time and under identical reopening scenarios. Under nominal probabilities, policies from the expected-loss and distributionally robust models yielded predicted total completion delays of 273.29 and 275.19 train-min, respectively. Allowing scenario probabilities to vary within the same range gave worst-case expected losses of 311.79 and 310.43 train-min, respectively. The distributionally robust policy accepted a slightly higher mean predicted delay to reduce potential losses when high-delay scenario probabilities were underestimated. This trade-off concerns schedule evaluation and selection; both models used identical reopening scenarios.

The ranking of the two models by actual delay changed with the closure-duration scale. When actual and predicted scales matched, mean total completion delay was 127.98 train-min for the distributionally robust model (Table~\ref{tab:robustness}). This was 3.11\% below the 132.09 train-min of the expected-loss model. At an actual scale of $0.5$, mean total completion delays were 100.92 and 97.64 train-min, respectively. At a scale of $2$, they were 284.37 and 255.42 train-min. The distributionally robust model had 3.36\% and 11.34\% higher mean delays at these scales. More cautious probability evaluation may therefore not offset systematic errors in closure-duration predictions. If actual durations repeatedly exceed or fall below predictions, field observations can first be used to recalibrate reopening-time scenarios. The delay reduction from protection against probability uncertainty can then be evaluated. Judging policies solely by lower worst-case predicted losses may overstate their benefits under closure-duration prediction errors.

\FloatBarrier

\section{Conclusions}
We develop a distributionally robust model for in-station train rescheduling under resource closures, arrival delays, and processing-time extensions. The model updates reopening-time predictions with field observations, and GNN-LBBD coordinates routing and resource scheduling. On the test instances, GNN-LBBD reduced mean total completion delay by 16.63\% relative to TS and 20.06\% relative to VNS. Its mean total computation time was close to that of TS. Ablation experiments further showed that graph learning-assisted ordering reduced total delay after resource lower bounds and cut strengthening had been enabled. The solution order of resource relaxations is thus an effective component for improving schedule quality within a limited computation budget.

Field observations of the resolution stage change train holding and routing during closures by revising reopening-time estimates. In the information comparisons, using stage observations reduced mean total completion delay by approximately 21.7\%. Execution records showed that track reassignments allowed trains to complete their operations on alternative routes before reopening. Earlier release of tracks also enabled subsequent arrivals and reduced on-track waiting for following trains. The benefits therefore extended beyond trains originally assigned to the closed track.

Distributionally robust evaluation balances mean predicted delay against losses under adverse probability distributions. When actual and predicted duration scales matched, it reduced mean total completion delay by 3.11\% relative to the expected-loss model. Under systematic duration prediction errors, a more cautious probability evaluation may not yield lower actual delays. Scale experiments also showed that estimating benefits from reductions in disturbance inputs may overstate the corresponding delay improvement. When closure durations are consistently overestimated or underestimated, field observations can first be used to recalibrate reopening-time scenarios. The operational benefits of protection against probability uncertainty can then be evaluated. Future work could jointly calibrate reopening-time scenarios and probability ambiguity sets, then assess scheduling performance across station layouts and operating loads.

\section*{Code Availability}
The source code and README files used in this study are publicly available at:\url{https://github.com/yuncifor/RESCHEDULING-UNDER-DISRUPTIONS}

\bibliographystyle{unsrtnat}

\bibliography{references}

\end{document}